\documentclass{article}
\PassOptionsToPackage{numbers}{natbib}

\usepackage[final]{neurips_2021}

\usepackage{algorithm, algorithmic}
\usepackage{amsmath}
\usepackage{subfigure}
\usepackage{amssymb}

\usepackage{graphicx}
\usepackage{wrapfig}
\usepackage{makecell,rotating,multirow,diagbox}
\usepackage{bbm}
\usepackage{wrapfig}

\usepackage[utf8]{inputenc} 
\usepackage[T1]{fontenc}    
\usepackage{hyperref}       
\usepackage{url}            
\usepackage{booktabs}       
\usepackage{amsfonts}       
\usepackage{nicefrac}       
\usepackage{microtype}      
\usepackage{xcolor}         
\usepackage{multirow}

\usepackage{bbm}

\begin{document}

\title{Exploring the Diversity and Invariance in Yourself for Visual Pre-Training Task}

\author{Longhui Wei\textsuperscript{1,2}, Lingxi Xie\textsuperscript{2}, Wengang Zhou\textsuperscript{1}, Houqiang Li\textsuperscript{1}, Qi Tian\textsuperscript{2}\\
\textsuperscript{1}University of Science and Technology of China,\quad\textsuperscript{2}Huawei Inc.\\
\texttt{weilh2568@gmail.com}, \texttt{198808xc@gmail.com}\\
\texttt{zhwg@ustc.edu.cn}, \texttt{lihq@ustc.edu.cn}, \texttt{tian.qi1@huawei.com}}

\maketitle

\begin{abstract}
Recently, self-supervised learning methods have achieved remarkable success in visual pre-training task. By simply pulling the different augmented views of each image together or other novel mechanisms, they can learn much unsupervised knowledge and significantly improve the transfer performance of pre-training models.
However, these works still cannot avoid the representation collapse problem,~\emph{i.e.}, they only focus on limited regions or the extracted features on totally different regions inside each image are nearly the same. Generally, this problem makes the pre-training models cannot sufficiently describe the multi-grained information inside images, which further limits 
the upper bound of their transfer performance. 
To alleviate this issue, this paper introduces a simple but effective mechanism, called \textbf{E}xploring the \textbf{D}iversity and \textbf{I}nvariance in \textbf{Y}ourself (E-DIY). 
By simply pushing the most different regions inside each augmented view away, E-DIY can preserve the diversity of extracted region-level features. By pulling the most similar regions from different augmented views of the same image together, E-DIY can ensure the robustness of region-level features. Benefited from the above diversity and invariance exploring mechanism, E-DIY maximally extracts the multi-grained visual information inside each image. Extensive experiments on downstream tasks demonstrate the superiority of our proposed approach,~\emph{e.g.}, there are $2.1\%$ improvements compared with the strong baseline BYOL on COCO while fine-tuning Mask R-CNN with the R50-C4 backbone and $1 \times$ learning schedule. 
\end{abstract}

\section{Introduction}

Nowadays, contrastive learning based methods~\cite{he2020momentum,chen2020simple,chen2020improved,tian2019contrastive,wei2020can} have achieved remarkable success in computer vision tasks. With simply pulling the different augmented views of each image together and pushing the different images away, they can learn much unsupervised knowledge from large-scale unlabeled images. Extensive experiments have shown these works can achieve competitive or sometimes even better results on downstream tasks compared with the fully-supervised learning methods~\cite{cubuk2018autoaugment,cubuk2020randaugment,zhang2017mixup,khosla2020supervised}. For the above appealing points taken by these self-supervised learning methods, more and more researchers pay attention on this to learn good visual pre-training models.

Recently, there are mainly two new directions for self-supervised learning methods. One is to weaken the necessity of negative pairs for avoiding careful treatment of these pairs~\cite{grill2020bootstrap,xie2020propagate}. For example, BYOL~\cite{grill2020bootstrap}, the first work with only pulling the different augmented views of each image together, can achieve much competitive transfer performances on downstream tasks. 
The other is to focus on region-level (or pixel-level) information but not image-level as usual~\cite{wang2020dense,xie2020propagate},~\emph{e.g.}, Xie~\emph{et al.}~\cite{xie2020propagate} proposed Pixel-to-Propagation (PixPro) module to explore the region-level consistency and extract discriminative region-level features.
\begin{wrapfigure}{r}{10cm}
    \centering
    \includegraphics[width=0.7\textwidth]{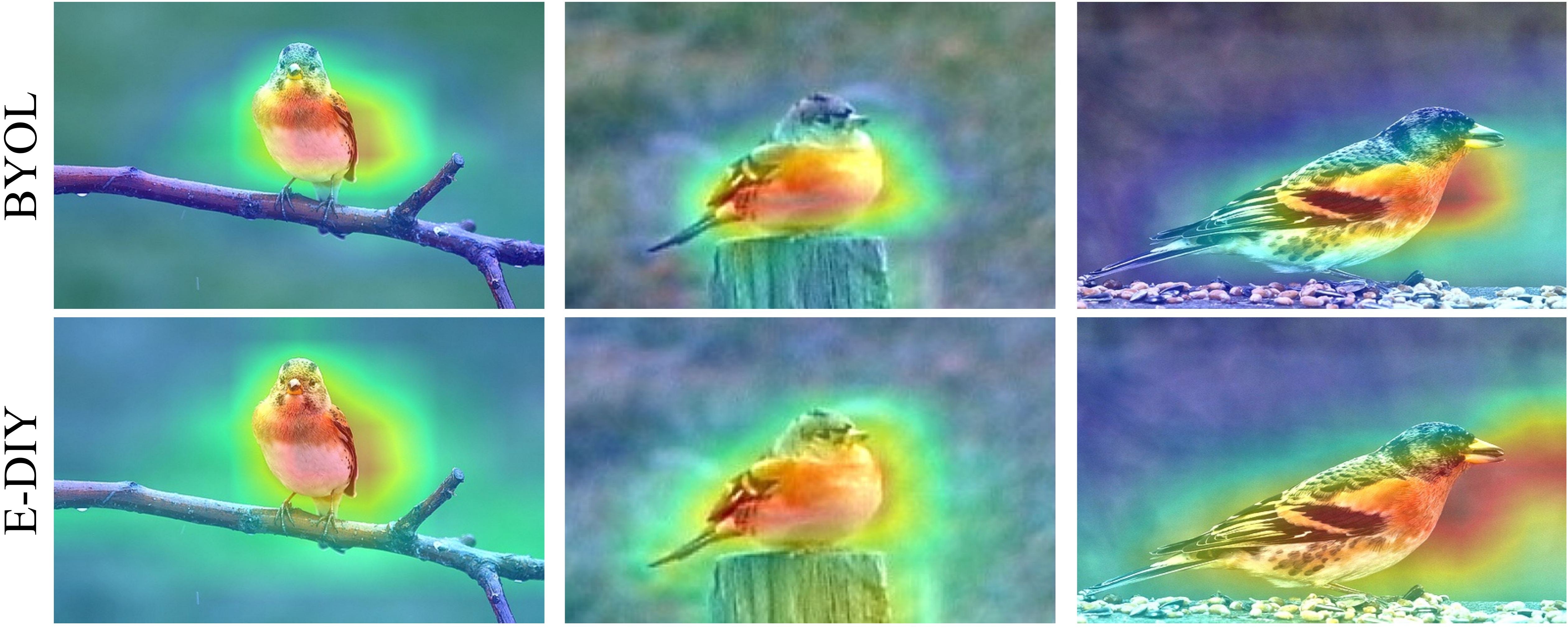}
    \caption{Illustration of the saliency maps generated by different pre-training methods. Generally, BYOL tends to attend on limited regions, but our proposed E-DIY tends to learn pre-training knowledge from more regions.}
    \label{fig:intro}
\end{wrapfigure}
Though extensive experiments show the above works can significantly push forward the transfer performance of pre-training models, they still meet the representation collapse challenge and thus cannot describe the whole information inside each image.

For example, as shown in Fig.~\ref{fig:intro}, because BYOL only targets to pull each different augmented views together, the learned knowledge is biased inevitably and focuses on limited regions (\emph{e.g.}, only attending on the bird body), which is somewhat similar with the character of the supervised learning methods~\cite{khosla2020supervised,wei2020circumventing}. Moreover, the region-level self-supervised learning methods, taking PixPro as an example, can force the pre-training models to focus on the region-level information. However, it only preserves the consistency of the features extracted on neighborhood regions, but cannot ensure the diversity of features extracted on totally different regions inside each image. Therefore, the useful learned knowledge can also be restricted to limited regions (\emph{e.g.}, only the overlapped regions of two augmented views). We call the above phenomenon as the representation collapse problem. Generally, a perfect pre-training model should contain multi-grained information (\emph{e.g.}, the bird, the stump and the background) as much as possible. Therefore, though the above works have achieved good transfer performances on downstream tasks, there is still much room for further improvement.



Target to address this issue, we propose a simple but effective mechanism, called \textbf{E}xploring the \textbf{D}iversity and \textbf{I}nvariance in \textbf{Y}ourself (E-DIY). Based on BYOL, E-DIY additionally includes two modules, the region-level diversity exploring module (R-DEM) for 
enhancing the diversity of extracted features on totally different regions inside each image and the region-level invariance exploring module (R-IEM) for ensuring the robustness of extracted region-level features. As shown in Fig.~\ref{fig:fig2}, the most similar regions in two augmented views will be found and pulled together in R-IEM, which can enhance the region-level invariance compared with only pulling the global features in BYOL. Meantime, the most dissimilar regions inside each augmented view will be searched and pushed away in R-DEM, which can guide the model to learn on more regions other than limited biased discriminative regions (\emph{e.g.}, the regions only benefitting to the classification task). Therefore, compared with BYOL only ensuring the image-level invariance and PixPro aiming for the region-level consistency, E-DIY targets to explore both the diversity and invariance inside each image. Generally, E-DIY can maximally learn fine-grained information on each region and meantime ensure the robustness of extracted features, which benefits a lot to visual pre-training task.

We evaluate E-DIY on various detection and segmentation downstream tasks. Extensive experimental results show E-DIY heavily improves the transfer performance of pre-training models. For example, compared with the strong baseline BYOL, there are $2.1\%$/$1.7\%$ improvements on COCO detection/instance segmentation task while fine-tuning Mask R-CNN with R50-C4 backbone and $1 \times$ learning schedule. Moreover, compared with the recent region-level pre-training works~\cite{wang2020dense,xie2020propagate}, our proposed E-DIY still shows good advantages on downstream tasks, which further reflects the necessity of exploring both the region-level invariance and diversity in visual pre-training task. More details of E-DIY and discussions with previous works are shown in Sec.~\ref{sec:3}.

\section{Related Work}
The key for self-supervised learning~\cite{he2020momentum,zhuang2019local,gutmann2010noise,doersch2015unsupervised,wei2020co2,PCL,wang2021contrastive,gidaris2020online,bhunia2021vectorization,misra2020self,han2020self,kong2020cycle,jabri2020space,wang2019self} is how to design the pretext task. Previously, the commonly used pretext tasks in self-supervised learning field are jigsaw puzzle~\cite{wei2019iterative,doersch2015unsupervised}, image restoration~\cite{kingma2013auto,rezende2014stochastic}, image rotation prediction\cite{gidaris2018unsupervised,chen2019self} and \textit{etc}. Though these works indeed have the ability to extract useful visual information from massive unlabeled images, there is still much room to achieve competitive performance as the supervised-learning methods~\cite{cubuk2018autoaugment,cubuk2020randaugment,khosla2020supervised,yun2019cutmix}. 

Recently, contrastive learning sheds a new way to effectively learn unsupervised representation. Compared with the previous pretext tasks, contrastive learning heavily pushes forward the upper-bound performance of self-supervised learning methods. Inspired by the above, lots of contrastive learning based works~\cite{wei2020can,he2020momentum,chen2020improved,chen2020simple,grill2020bootstrap,wu2018unsupervised} are designed. For example, Momentum Contrast (MoCo)~\cite{he2020momentum} proposed to build a dynamic memory bank for enlarging the contrastive negative samples, which can break the limitation of GPU memory and heavily improve the final performance. Compared with the traditional supervised learning methods, MoCo firstly revealed the advantages of self-supervised learning methods in the visual pre-training task. Moreover, SimCLR~\cite{chen2020simple} evaluated the importance of data augmentation for contrastive learning based self-supervised methods, and significantly improved the transfer performance. Additionally, InfoMin~\cite{tian2020makes} further analyzed the necessity of view selection. 
Other than the contrastive learning based methods, some other novel approaches have been also proposed. For example, BYOL~\cite{grill2020bootstrap} evaluated that simply pulling the different augmented views of each image together can also achieve competitive performance. SwAV~\cite{caron2020} proposed one online-clustering based mechanism to further improve the performance of pre-training models.

However, most of the above works only consider the instance-level discrimination, but ignore the details inside each image. Target to achieve better results on dense prediction downstream tasks,~\emph{e.g.}, detection and segmentation task. Dense Contrast Learning (DenseCL)~\cite{wang2020dense} firstly utilized the region-level (or pixel-level) contrast loss to enhance the region-level discrimination of pre-training models. Besides, Pixel-to-Propagation (PixPro)~\cite{xie2020propagate} proposed to explore the region-level consistency and utilized the pixel propagation module to achieve spatial smoothness of region-level representation. By simply adding the region-level contrast or consistency loss, the above two works significantly improve the transfer performance of pre-training models,~\emph{e.g.}, compared with the baseline (MoCo), there are $0.5\%$ improvements for DenseCL on COCO detection task.

Our work is also a region-level based self-supervised method. Different from DenseCL and PixPro, the core of E-DIY is to explore the naturally existing diversity inside each image (its importance is ignored previously). Meanwhile, target to ensure the robustness of extracted region-level features, E-DIY also utilizes the region-level invariance exploring mechanism. More details about E-DIY are described in the next section.

\section{Explore the Region-level Diversity and Invariance in Yourself}
\label{sec:3}
\subsection{Preliminaris: Self-supervised Learning}

The core of self-supervised learning is to design a suitable pretext task for maximally guiding the network to learn useful knowledge on unlabeled data. Suppose the unlabeled data as $D$ and pretext task as $P$, the optimization goal of self-supervised learning can  be represented as:
\begin{equation}
    \mathrm{Knowledge} = \mathop{\arg\max}_{P}\mathrm{Optim}(P,D),
\end{equation}
where $\mathrm{Optim}$ denotes the optimizer for unsupervised knowledge learning.


Bootstrap Your Own Latent (BYOL)~\cite{grill2020bootstrap} is a current state-of-the-art self-supervised learning method. Through a simple pretext task with only keeping the instance-level representation consistency of two different augmented views of each image, BYOL can effectively learn discriminative unsupervised representation. Suppose one image $\mathbf{x}_i$, its two augmented views can be represented as $\mathbf{x}_i^{\dag}$ and $\mathbf{x}_i^{\ddag}$, respectively. The online and target networks for different views are represented as $\mathbf{f}$ and $\mathbf{g}$, thus the loss of BYOL can be formulated as:
\begin{equation}\label{eq:byol}
    \mathcal{L}^\mathrm{BYOL}_{\mathbf{x_i}} = 4-2\frac{\mathbf{f}( \mathbf{x}_i^{\dag})^\top\cdot\mathbf{g}(\mathbf{x}_i^{\ddag})}{||\mathbf{f}( \mathbf{x}_i^{\dag})||_{2}\cdot||\mathbf{g}(\mathbf{x}_i^{\ddag})||_{2}}-2\frac{\mathbf{g}( \mathbf{x}_i^{\dag})^\top\cdot\mathbf{f}(\mathbf{x}_i^{\ddag})}{||\mathbf{g}( \mathbf{x}_i^{\dag})||_{2}\cdot||\mathbf{f}(\mathbf{x}_i^{\ddag})||_{2}},
\end{equation}
The core of BYOL is to prevent the model collapse by designing a prediction task with an iteratively enhanced target network and online network. However, though BYOL is relatively simple compared with contrastive learning based methods~\cite{he2020momentum,chen2020improved}, it still meets the overfitting challenge.

As shown in Eq.~\ref{eq:byol}, suppose the augmentation policies for online network $\mathbf{f}$ and target network $\mathbf{g}$ are the same, it is hard to optimize for $\mathbf{f}$ because that any random weights but same with $\mathbf{g}$ can make the loss minimum. Therefore, the learned knowledge is random and meaningless. Target to address this problem, researchers mainly carefully design the augmentation policy. By regularizing the input with different augmentation policies for online network $\mathbf{f}$ and target network $\mathbf{g}$, it actually can prevent the model collapse but further cause the representation collapse problem. For example, while two augmented views are cropped from totally different regions in one image, the optimization goal of Eq.~\ref{eq:byol} is to pull their generated features together, which is unreasonable (\emph{e.g.}, the feature generated on the sky region (the background) should not be the same with the feature generated on the bird region (the foreground)). However, while there are overlapped regions for these two inputs, the optimization goal will guide the model to focus on the overlapped regions but ignore other information. As shown in Fig.~\ref{fig:intro}, BYOL only attends on limited discriminative regions (\emph{i.e.}, part of the foreground), which restricts its upper-bound performance on downstream tasks. The above phenomenon reveals the representation collapse problem, for that the pre-training models learned by the above scheme cannot fully describe the multi-grained information inside each image. 
Given the suppose that, more knowledge learned in the pre-training stage, better performance on downstream tasks, the naturally existing multi-grained information inside every single image should be further exploited for better pre-training models. 

\begin{figure}[!t]
    \centering
    \includegraphics[width=1.0\textwidth]{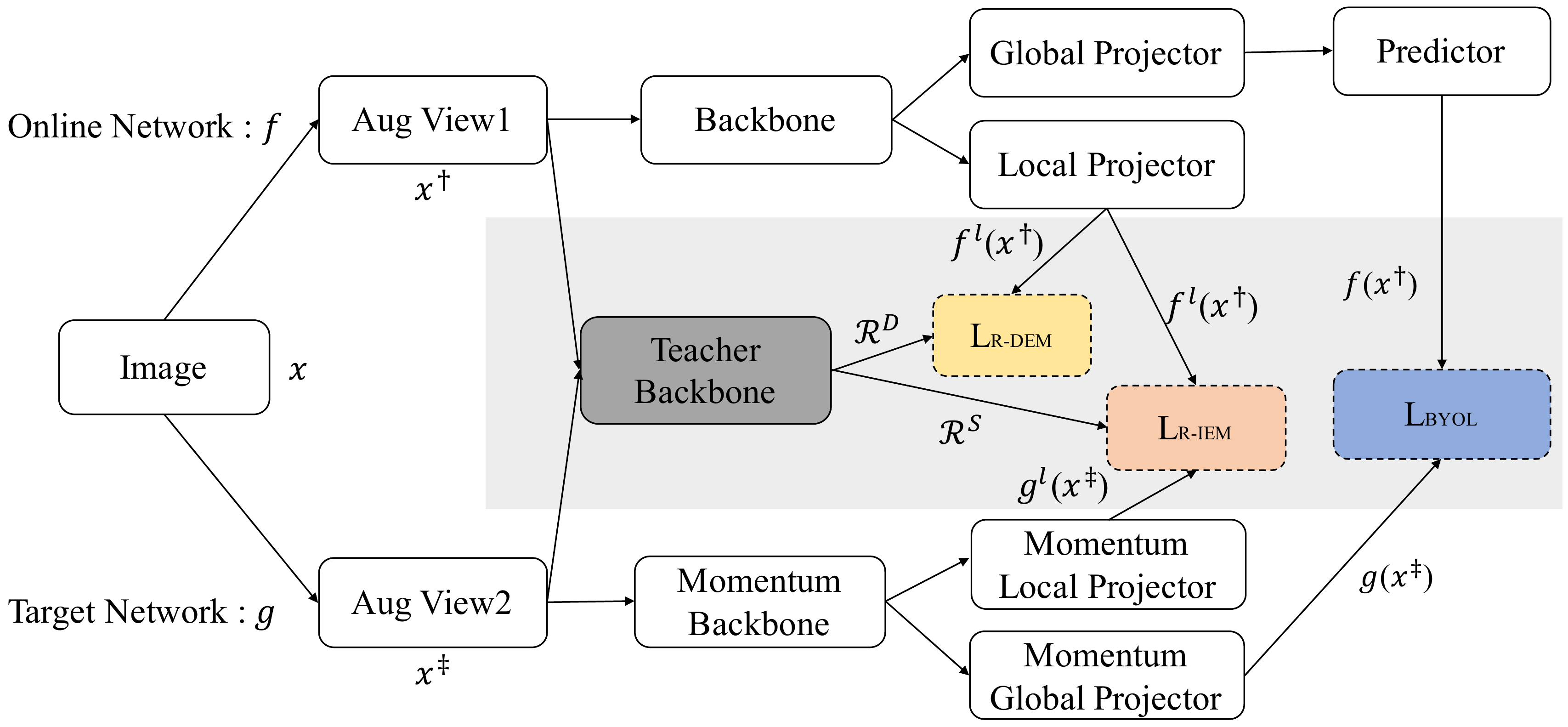}
    \caption{The framework of E-DIY. Teacher Backbone represents a pre-trained teacher model for computing the region similarity and then guiding the online network to learn the region-level relationship (diversity and invariance character inside each image). Global Projector is to output the instance-level feature and Local Projector is to output the region-level feature. The architectures of these two modules are the same as the projector in BYOL~\cite{grill2020bootstrap}. More details about E-DIY can be seen in Sec.~\ref{sec:3}.}
    \label{fig:fig1}
\end{figure}

\subsection{Region-level Diversity and Invariance Exploring Mechanism}
\begin{figure}[!t]
    \centering
    \includegraphics[width=1.0\textwidth]{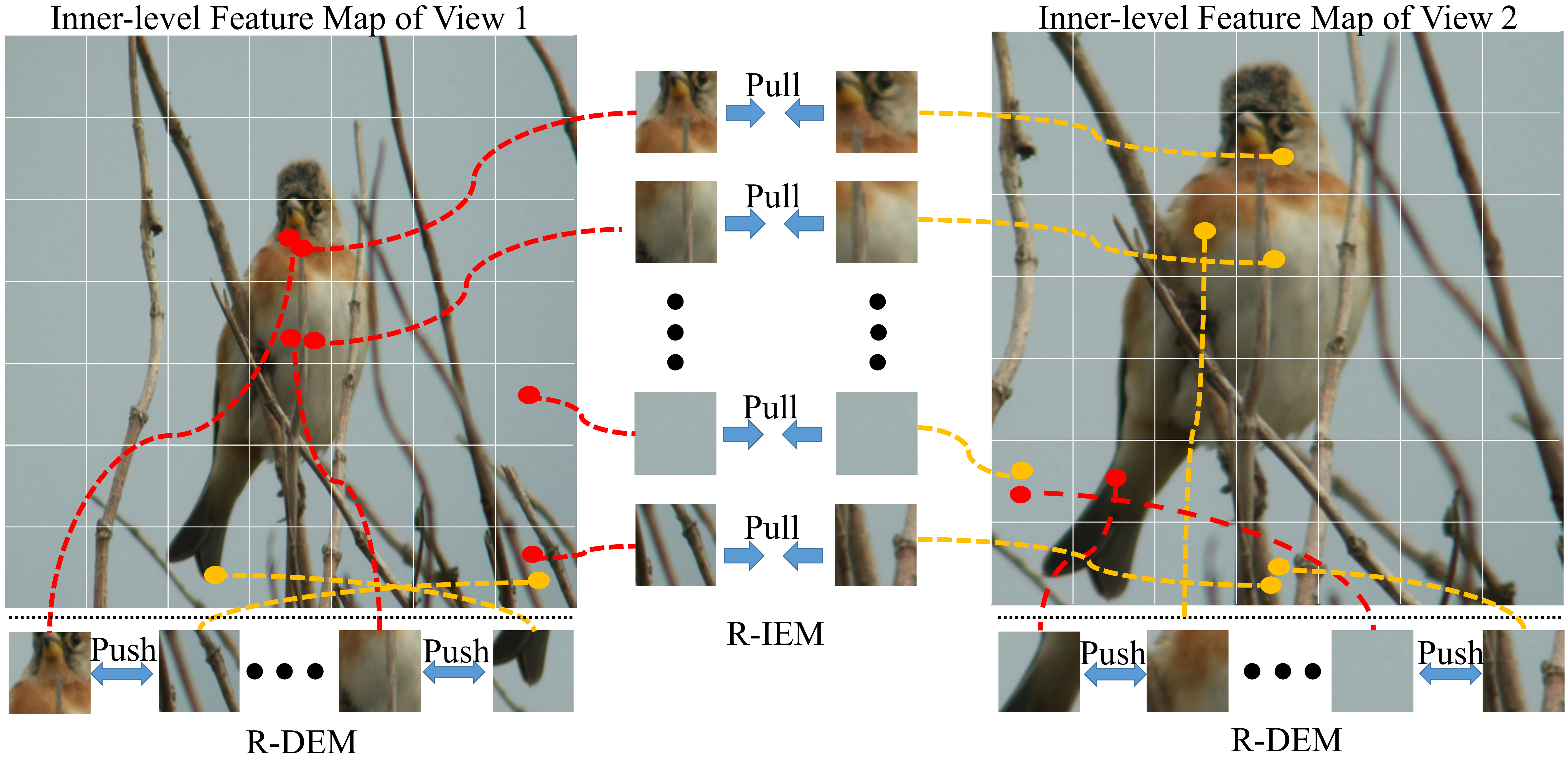}
    \caption{Illustration of the proposed region-level diversity exploring module (R-DEM) and region-level invariance exploring module (R-IEM). Generally, R-DEM tries to push the most different regions inside the same augmented view away, and R-IEM tries to pull the most similar regions among the different augmented views of the same image together. The matched pairs (different or similar regions) are searched by ranking the similarity of their region features. More details can be seen in Sec.~\ref{sec:3}.}
    \label{fig:fig2}
\end{figure}

Target to enhance the transfer performance of BYOL, we propose a region-level diversity and invariance exploring mechanism (called E-DIY). The overall framework of this mechanism is illustrated in Fig.~\ref{fig:fig2}. E-DIY is motivated by the information inside each image is multi-grained. As shown in Fig.~\ref{fig:intro}, even for the simple image with one bird, it still contains two regions, ~\emph{i.e.}, the background and foreground. Moreover, in the foreground, its head region in the bird is different from its body region. Therefore, it is unreasonable to describe the image with limited discriminative region features for the visual pre-training task. Target to address this issue, we firstly propose the region-level diversity exploring module (R-DEM) to enhance the richness of extracted features. As shown in Fig.~\ref{fig:fig2}, R-DEM seeks the most dissimilar regions in one image, and tries to push them away in feature space. Obviously, R-DEM aims for exploring the naturally existing diversity character inside each image. However, only pushing the dissimilar regions tends to make the learned features sensitive to spatial relation, which is harmful to the pre-training task. To further ensure the robustness of pre-training model, we propose the region-level invariance exploring module (R-IEM). Similar to R-DEM, R-IEM seeks the most similar regions in different augmented views, and tries to pull them together in the latent feature space. Thus, guided by R-IEM, the pre-training model can be ensured to learn robust and discriminative region-level features.


Based on the above discussion, the loss function of our E-DIY for each image $\mathbf{x}_i$ can be formulated as:

\begin{equation}\label{eq:1}
    \mathcal{L}^\mathrm{E-DIY}_{\mathbf{x}_i} = \lambda_{1}\mathcal{L}^\mathrm{R-DEM}_{\mathbf{x}_i}+\lambda_{2}\mathcal{L}^\mathrm{R-IEM}_{\mathbf{x}_i}+\lambda_{3}\mathcal{L}^\mathrm{BYOL}_{\mathbf{x}_i},
\end{equation}
where $\lambda$ represents the loss weight for each module. The details of R-DEM and R-IEM are described as follows. 

\subsubsection{Region-level Diversity Exploring Module (R-DEM)}
The motivation of our Region-level Diversity Exploring Module (R-DEM) is to prevent the overfitting problem and extract the multi-grained information as much as possible. Therefore, the core of R-DEM is how to prevent the representation collapse (~\emph{i.e.}, avoid focusing on limited regions). Target to achieve this, R-DEM extends the representation space of images by enlarging the distance of different regions inside each image. However, it is impossible to require the region labels for every image because of the costly annotations. To relieve this burden, and naturally combine with the self-supervised learning framework, R-DEM utilizes one extra self-supervised learning model (~\emph{e.g.}, the model trained by BYOL) as the teacher to seek the different regions by the feature similarity degree. Therefore, given the region $\mathbf{x}_{i:{\mathbf{r}_j}}^{\dag}$ in one augmented view of image $\mathbf{x}_{i}$, its most different region inside the same augmented view can be found by the following:
\begin{equation}\label{eq:1}
    \mathcal{R}_{\mathbf{x}_{i:{\mathbf{r}_j}}^{\dag}}^\mathbf{D} = \mathop{\arg\min}_{z}sim(\mathbf{T}({\mathbf{x}_{i}^{\dag}})_{\mathbf{r}_z},\mathbf{T}({\mathbf{x}_{i}^{\dag}})_{\mathbf{r}_j}),
\end{equation}
where $\mathbf{T}$ represents the feature encoder of teacher model,
$\mathbf{T}({\mathbf{x}_{i}^{\dag}})_{\mathbf{r}_z}$ represents the feature of one region $\mathbf{x}_{i:\mathbf{r}_z}^\dag$ in the augmented view $\mathbf{x}_{i}^\dag$.
After finding the most different region for each region in one image, the loss function of D-REM can be formulated as:
\begin{equation}\label{eq:1}
    \mathcal{L}^\mathrm{R-DEM}_{\mathbf{x}_i} = \frac{1}{n} \sum_{j=1}^n{\frac{\mathbf{f}^{l}(\mathbf{x}_{i}^{\dag})_{\mathbf{r}_j}^\top\cdot\mathbf{f}^{l}(\mathbf{x}_{i}^{\dag})_{\mathcal{R}_{\mathbf{x}_{i:{\mathbf{r}_j}}^{\dag}}^\mathbf{D}}}{||\mathbf{f}^{l}(\mathbf{x}_{i}^{\dag})_{\mathbf{r}_j}||_{2}\cdot||\mathbf{f}^{l}(\mathbf{x}_{i}^{\dag})_{\mathcal{R}_{\mathbf{x}_{i:{\mathbf{r}_j}}^{\dag}}^\mathbf{D}}||_{2}}}+\frac{1}{n} \sum_{j=1}^n{\frac{\mathbf{f}^{l}(\mathbf{x}_{i}^{\ddag})_{\mathbf{r}_j}^\top\cdot\mathbf{f}^{l}(\mathbf{x}_{i}^{\ddag})_{\mathcal{R}_{\mathbf{x}_{i:{\mathbf{r}_j}}^{\ddag}}^\mathbf{D}}}{||\mathbf{f}^{l}(\mathbf{x}_{i}^{\ddag})_{\mathbf{r}_j}||_{2}\cdot||\mathbf{f}^{l}(\mathbf{x}_{i}^{\ddag})_{\mathcal{R}_{\mathbf{x}_{i:{\mathbf{r}_j}}^{\ddag}}^\mathbf{D}}||_{2}}},
\end{equation}
where $n$ denotes the sum of region number in each augmented view, $\mathbf{f}^l$ represents the region-level feature encoder (as shown in Fig.~\ref{fig:fig1}), thus $\mathbf{f}^{l}(\mathbf{x}_{i}^{\dag})_{\mathbf{r}_j}$ denotes the feature of the region $\mathbf{x}_{i:{\mathbf{r}_j}}^{\dag}$ in the feature map $\mathbf{f}^l(\mathbf{x}_{i}^{\dag})$.

\subsubsection{Region-level Invariance Exploring Module (R-IEM)}
Contrary to R-DEM, Region-level Invariance Exploring Module (R-IEM) targets to exploit the naturally existing region-level consistency in different augmented views of the same image. Thus, the core of R-IEM is to look for consistent regions. Similar to R-DEM, R-IEM also utilizes the teacher guided feature similarity ranking scheme to find the most matched region of each query region:
\begin{equation}\label{eq:1}
    \mathcal{R}_{\mathbf{x}_{i:{\mathbf{r}_j}}^{\dag}}^\mathbf{S} = \mathop{\arg\max}_{z}sim(\mathbf{T}({\mathbf{x}_{i}^{\ddag}})_{\mathbf{r}_z},\mathbf{T}({\mathbf{x}_{i}^{\dag}})_{\mathbf{r}_j}),
\end{equation}
Therefore, the loss function of R-IEM can be represented as:
\begin{equation}\label{eq:1}
    \mathcal{L}^\mathrm{R-IEM}_{\mathbf{x}_i} = 2 - \frac{1}{n} \sum_{j=1}^n{\frac{\mathbf{f}^{l}(\mathbf{x}_{i}^{\dag})_{{\mathbf{r}_j}}^\top\cdot\mathbf{g}^{l}(\mathbf{x}_{i}^{\ddag})_{\mathcal{R}_{\mathbf{x}_{i:{\mathbf{r}_j}}^{\dag}}^\mathbf{S}}}{||\mathbf{f}^{l}(\mathbf{x}_{i}^{\dag})_{{\mathbf{r}_j}}||_{2}\cdot||\mathbf{g}^{l}(\mathbf{x}_{i}^{\ddag})_{\mathcal{R}_{\mathbf{x}_{i:{\mathbf{r}_j}}^{\dag}}^\mathbf{S}}||_{2}}}-\frac{1}{n} \sum_{j=1}^n{\frac{\mathbf{f}^{l}(\mathbf{x}_{i}^{\ddag})_{{\mathbf{r}_j}}^\top\cdot\mathbf{g}^{l}(\mathbf{x}_{i}^{\dag})_{\mathcal{R}_{\mathbf{x}_{i:{\mathbf{r}_j}}^{\ddag}}^\mathbf{S}}}{||\mathbf{f}^{l}(\mathbf{x}_{i}^{\ddag})_{{\mathbf{r}_j}}||_{2}\cdot||\mathbf{g}^{l}(\mathbf{x}_{i}^{\dag})_{\mathcal{R}_{\mathbf{x}_{i:{\mathbf{r}_j}}^{\ddag}}^\mathbf{S}}||_{2}}},
\end{equation}
where $\mathbf{g}^l$ represents the region-level feature encoder of the target network.

\subsection{Difference from Previous Works}
The major difference with previous region-level based works~\cite{xie2020propagate,wang2020dense} is that our E-DIY is designed to explore the multi-grained information inside every single image. Unlike the PixPro only keeping the region-level consistency of two augmented views, or the DenseCL utilizing other global images in the batch as the negative samples, E-DIY looks for the most different regions inside each image and enlarges the similarity of their features for maximally guiding the pre-training model to attend on multiple regions. Moreover, to ensure that the extracted region-level features are discriminative and robust, E-DIY further utilizes the region-level consistency principle, which is similar to PixPro. Though the modification of E-DIY is simple compared with these previous works, extensive experiments on downstream tasks show E-DIY heavily pushes forward the transfer performances on various downstream tasks.

Moreover, to our best knowledge, E-DIY is the first to reveal that the representation collapse problem does harm to pre-training models. No matter for the supervised learning or self-supervised learning based pre-training methods, they generally overfit on their pretext task. As a result, the learned knowledge is coarse and contains only a small share of the overall knowledge provided by original data. E-DIY struggles with this challenge by enlarging the diversity of learned knowledge. Therefore, other than self-supervised learning methods (\emph{e.g.}, BYOL), E-DIY can also be regarded as a supplement to other supervised pre-training methods, which is not the core of our paper and will be left for future research.

\section{Experiments}
\subsection{Datasets}
ImageNet-1K~\cite{ImageNet} is the most commonly used pre-training dataset in computer vision community. It contains $1.28$ million images of $1,000$ classes. In this paper, we also pre-train E-DIY on ImageNet-1K and evaluate its effectiveness on various downstream datasets. 

Following DenseCL~\cite{wang2020dense}, we mainly conduct evaluations on PASCAL VOC~\cite{everingham2010pascal} and COCO~\cite{lin2014microsoft}. 
PASCAL VOC is a commonly used detection dataset, and the size of this dataset is relatively small. In the following, we denote PASCAL VOC as VOC for short. In this paper, all of the experiments on VOC will be trained on VOC$_{trainval07+12}$, and tested on VOC$_{test07}$. COCO is a more large-scale dataset, and it contains varies of annotations,~\emph{e.g.}, bounding box, instance segmentation, keypoint and \textit{}{etc}. In this paper, we mainly conduct evaluations on detection and instance segmentation tasks of COCO. All of the models will be trained on COCO$_{train2017}$ and tested on COCO$_{val2017}$. 

\subsection{Implementation Details}
In the pre-training stage while training E-DIY, we strictly follow the image augmentation strategy in BYOL~\cite{grill2020bootstrap},~\emph{i.e.}, random resized crop operation followed by color jitter, grayscale conversion, Gaussian blur, solarization, and random flip. For the network architecture, we also follow the same design of BYOL. Differently, there are two projectors in E-DIY, where one is for the instance-level feature and the other is for the region-level feature. The architecture of these two projectors is the same as the projector in BYOL.
Moreover, target to compute the region-level similarity, there is one additional teacher model (ResNet-50~\cite{he2016identity}, trained by BYOL). We calculate the similarity of regions with the last feature map in layer-4 of the teacher model. For the mini-batch of each GPU, we randomly sample $128$ images. Finally, we utilize $16$ GPUs, and thus a total of $2048$ images are sampled in each iteration. Following BYOL, we adopt LARS optimizer~\cite{you2017scaling} and the initial learning rate is set as $3.2$. To address the limited batch size and reproduce the results of BYOL, we utilize the gradient accumulation strategy. Finally, we train E-DIY with $1,000$ epochs on Imagenet-1K.


In the downstream evaluation stage, we utilize Detectron2~\cite{wu2019detectron2} framework to fine-tune the Faster R-CNN~\cite{ren2015faster} with R50-C4 backbone on VOC. For the training and testing on the detection task of VOC, we strictly follow the settings in MoCo~\cite{he2020momentum}. Additionally, for the detection and instance segmentation task on COCO, we fine-tune the Mask R-CNN~\cite{he2017mask} with different backbones and learning schedules as MoCo. More details can be seen in this paper~\cite{he2020momentum}.

\begin{table}
  \caption{Results of different pre-training approaches while fine-tuning Mask-RCNN with the R50-FPN backbone and $1 \times$ learning schedule on COCO. All of the models are pre-trained with 300 epochs on ImageNet-1K. TG represents our proposed teacher guided region sampling scheme, R denotes the random region sampling scheme, and R-INS represents that randomly sampling the global feature of another image as DenseCL~\cite{wang2020dense}. COCO$_{\mathrm{Det}}$ denotes the detection task on COCO, and COCO$_{\mathrm{InsSeg}}$ represents the instance segmentation task on COCO, respectively.}
  \label{Ablation study}
  \centering
  \begin{tabular}{lcc|ccc|ccc}
    \toprule
    \multirow{2}*{BYOL} & \multirow{2}*{R-DEM} & \multirow{2}*{R-IEM} & \multicolumn{3}{c|}{COCO$_{\mathrm{Det}}$} & \multicolumn{3}{c}{COCO$_{\mathrm{InsSeg}}$} \\
    \cline{4-9}
    & & & \multicolumn{1}{c}{AP$^\mathrm{bb}$} & \multicolumn{1}{c}{AP$_{50}^\mathrm{bb}$} & \multicolumn{1}{c|}{AP$_{75}^\mathrm{bb}$} & \multicolumn{1}{c}{AP$^\mathrm{mk}$} & \multicolumn{1}{c}{AP$_{50}^\mathrm{mk}$} & \multicolumn{1}{c}{AP$_{75}^\mathrm{mk}$}\\
    \hline
    \checkmark & - & - & $40.1$  & $61.6$ & $43.7$ & $36.7$ &$58.4$ & $39.4$     \\
    - & TG & - & $35.3$  & $54.7$ & $38.4$ & $32.3$ &$52.0$ & $34.5$     \\
    - & - & TG & $3.8$  & $7.1$ & $3.7$ & $3.7$ &$6.8$ & $3.6$     \\
    - & TG & TG & $39.8$  & $59.9$ & $43.5$ & $36.2$ &$57.0$ & $38.9$     \\
    \checkmark & TG  & - & $40.5$ &$61.3$ &$44.2$ & $36.8$ & $58.3$ & $39.5$  \\
    \checkmark & -  & TG & $40.3$ & $61.5$ & $44.1$ & $36.8$ & $58.4$ & $39.2$      \\
    \checkmark & TG  & TG & $\textbf{41.1}$ & $\textbf{62.2}$ &$\textbf{45.0}$ & $\textbf{37.3}$ & $\textbf{58.9}$ & $\textbf{40.1}$    \\
    \checkmark & R  & R &$40.4$ &$61.6$ &$44.2$ &$36.9$ &$58.7$ &$39.4$      \\
    \checkmark & R-INS  & TG &$40.3$ &$61.5$ &$43.8$ &$36.7$ &$58.4$ &$39.1$      \\    
    \bottomrule
  \end{tabular}
\end{table}

\subsection{Ablation Study}
\textbf{Importance of region-level diversity and invariance exploring module.} In this section, we firstly evaluate the importance of our proposed region-level diversity and invariance exploring mechanism. As shown in Tab.~\ref{Ablation study}, only utilizing the R-DEM or R-IEM, the pre-training models fail to transfer into downstream tasks. Especially, while only adopting R-IEM to guide the model training, it will meet the model collapse problem. However, while combining the R-DEM or R-IEM with BYOL, the transfer performance can be slightly improved. For example, there are $0.4\%$ improvements on the Average Precision (AP) metric of COCO detection task while combing BYOL with R-DEM module.
Moreover, it is interesting that, the improvements of E-DIY (consisting of R-DEM, R-IEM and BYOL) are more significant than each of the combinations of every two modules, which further reveals these three modules are complementary in enhancing the performance of pre-training models.

Additionally, compared with the method that pulling the randomly sampled regions of the same augmented views together and pushing the randomly sampled regions of different augmented views away, E-DIY can learn much more transferable knowledge. This result demonstrates the effectiveness of our teacher guided region sampling strategy. Moreover, compared with the negative pairs sampling mechanism in DenseCL~\cite{wang2020dense} (which randomly sampled other different images as the negatives pairs for the anchor patch in current image), E-DIY heavily surpasses its performance. 
The significant improvements can mainly own to two reasons: fixing the feature misalignment problem and avoiding the representation collapse challenge. The negative pairs sampling mechanism in DenseCL tries to push the region-level features and global-level features, which exists the misalignment problem. Differently, E-DIY tries to handle the relations of different regions. Therefore, the feature misalignment problem does not exist in E-DIY. Moreover, E-DIY explores the diversity inside each image at least, which can guide the pre-training model to attend on more regions as much as possible. However, previous works~\cite{wang2020dense,xie2020propagate} cannot make sure this and easily meet the overfitting problem.

\textbf{The impact of hyper-parameters.} We also conduct evaluations on the hyper-parameters of E-DIY. As shown in Tab.~\ref{hp}, while reducing the loss weights of each module, the transfer performance on downstream tasks will drop with different degrees. For example, while reducing the loss weight of R-IEM module from $1.0$ to $0.1$, the transfer performance on COCO detection task drops from $40.3\%$ to $38.8\%$ on AP metric. The above results further evaluate the importance of each module in E-DIY for the visual pre-training task.

\begin{table}
  \caption{Ablation study on the loss weights of each module in E-DIY. All of the models are trained with 100 epochs on ImageNet-1K.}
  \label{hp}
  \centering
  \begin{tabular}{lcc|ccc|ccc}
    \toprule
    \multirow{2}*{BYOL} & \multirow{2}*{R-DEM} & \multirow{2}*{R-IEM}    & \multicolumn{3}{c|}{COCO$_\mathrm{Det}$} & \multicolumn{3}{c}{COCO$_\mathrm{InsSeg}$} \\
    \cline{4-9}
    & & & \multicolumn{1}{c}{AP$^\mathrm{bb}$} & \multicolumn{1}{c}{AP$_{50}^\mathrm{bb}$} & \multicolumn{1}{c|}{AP$_{75}^\mathrm{bb}$} & \multicolumn{1}{c}{AP$^\mathrm{mk}$} & \multicolumn{1}{c}{AP$_{50}^\mathrm{mk}$} & \multicolumn{1}{c}{AP$_{75}^\mathrm{mk}$}\\
    \hline
    $0.1$ & $1.0$ & $1.0$ & $38.6$  & $58.5$ & $41.9$ & $35.1$ &$55.7$ & $37.7$     \\ 
    $1.0$ & $0.1$ & $1.0$ & $40.1$  & $61.2$ & $43.6$ & $36.6$ &$58.1$ & $39.2$     \\   
    $1.0$ & $1.0$ & $0.1$ & $38.8$  & $59.3$ & $42.4$ & $35.4$ &$56.1$ & $37.9$     \\ 
    $1.0$ & $1.0$ & $1.0$ & $40.3$  & $60.8$ & $44.3$ & $36.6$ &$57.6$ & $39.1$     \\   
    \bottomrule
  \end{tabular}
\end{table}

\subsection{Comparisons with Previous Works on Downstream Tasks}
\begin{wraptable}[]{r}{7cm} 
  \caption{Results of different pre-training approaches on VOC detection task. Following BYOL, E-DIY is trained with 1000 epochs on ImageNet-1K. The reported numbers of E-DIY are the average of 3 trials.}
  \label{Results on VOC}
  \centering
  \begin{tabular}{l|ccc}
    \toprule
    \multirow{2}*{Approaches}   & \multicolumn{3}{c}{VOC} \\
    \cline{2-4}
    & \multicolumn{1}{c}{AP} & \multicolumn{1}{c}{AP$_{50}$} & \multicolumn{1}{c}{AP$_{75}$} \\
    \hline
    SimCLR~\cite{chen2020simple} & $51.5$ & $79.4$  & $55.6$   \\
    MoCo-v2~\cite{chen2020improved}  & $57.4$ & $82.5$ &$64.0$ \\
    Infomin~\cite{tian2020makes}  & $57.5$ & $82.5$ &$64.0$ \\
    SwAV~\cite{caron2020}  & $56.1$ & $82.6$ & $62.7$  \\
    SimSiam~\cite{chen2021}  & $57.0$ & $82.4$ & $63.7$  \\
    BYOL~\cite{grill2020bootstrap}  & $53.1$ & $81.9$ & $58.6$  \\
    \hline
    E-DIY  & $\textbf{58.4}$ & $\textbf{83.8}$ & $\textbf{65.7}$\\
    \bottomrule
  \end{tabular}
\end{wraptable}
\textbf{Evaluations on VOC.} We firstly conduct comparisons with previous works on VOC detection task. 
As shown in Tab.~\ref{Results on VOC}, E-DIY achieves competitive performance compared with recent state-of-the-art methods. Though the baseline (\emph{i.e.}, BYOL) achieves much worse performance on VOC, E-DIY can heavily push forward its performance (\emph{e.g}, there are $7.1\%$ improvements on AP$_{75}$ metric compared with BYOL). Moreover, compared with other recent state-of-the-art methods, E-DIY still shows competitive transfer performances. Generally, the above results demonstrate the advantages of our proposed method in visual pre-training task.

\textbf{Evaluations on COCO.} To further evaluate the effectiveness of E-DIY, we conduct evaluations on COCO detection and instance segmentation tasks. Following the experimental settings in MoCo~\cite{he2020momentum}, we fine-tune the Mask R-CNN with different backbones and learning schedules on COCO. All of the results are shown in Tab.~\ref{tab:2} and Tab.~\ref{tab:3}. As shown in these tables, E-DIY shows clear advantages on transferring the pre-trained knowledge. For example, while fine-tuning the Mask R-CNN with R50-C4 backbone and $2\times$ learning schedule on COCO, E-DIY achieves $42.4\%$ in AP metric on detection task, which surpasses BYOL with $1.2\%$ improvements. In the instance segmentation task, E-DIY also achieves $0.9\%$ improvements compared with BYOL. Moreover, compared with recent state-of-the-art methods, E-DIY also shows consistent advantages on different downstream tasks, which clearly demonstrates the effectiveness of E-DIY for visual pre-training task.

\begin{table*}[htb]
    \centering
    \caption{Results of adopting different pre-training models to fine-tune Mask R-CNN with the R50-FPN backbone and different learning schedules on COCO. Following BYOL, E-DIY is trained with 1000 epochs on ImageNet-1K. Most of the reported methods follow the fine-tuning protocol in MoCo, but ${\sharp}$ represents that the method follows the fine-tuning protocol in InfoMin.}
    \scalebox{0.75}{
    \begin{tabular}{l|ccc|ccc||ccc|ccc}
        \hline
        \multirow{3}{*}{Methods}  &  \multicolumn{6}{c||}{COCO$_\mathrm{Det}$} & \multicolumn{6}{c}{ COCO$_\mathrm{InsSeg}$} \\
        \cline{2-13}
        {} &  \multicolumn{3}{c|}{1$\times$ schedule} & \multicolumn{3}{c||}{2$\times$ schedule} & \multicolumn{3}{c|}{1$\times$ schedule}  & \multicolumn{3}{c}{2$\times$ schedule} \\
        \cline{2-13}
        {} &  \multicolumn{1}{c}{AP$^\mathrm{bb}$} & \multicolumn{1}{c}{AP$_{50}^\mathrm{bb}$} & \multicolumn{1}{c|}{AP$_{75}^\mathrm{bb}$} &
        \multicolumn{1}{c}{AP$^\mathrm{bb}$} & \multicolumn{1}{c}{AP$_{50}^\mathrm{bb}$} & \multicolumn{1}{c||}{AP$_{75}^\mathrm{bb}$} &
        \multicolumn{1}{c}{AP$^\mathrm{mk}$} & \multicolumn{1}{c}{AP$_{50}^\mathrm{mk}$} & \multicolumn{1}{c|}{AP$_{75}^\mathrm{mk}$} &
        \multicolumn{1}{c}{AP$^\mathrm{mk}$} & \multicolumn{1}{c}{AP$_{50}^\mathrm{mk}$} & \multicolumn{1}{c}{AP$_{75}^\mathrm{mk}$} \\
        \hline \hline
        FSup-Softmax~\cite{he2020momentum}&  38.9 & 59.6& 42.7& 40.6& 61.3& 44.4& 35.4& 56.5& 38.1& 36.8& 58.1& 39.5\\
        \hline
        MoCo-v2~\cite{chen2020improved}& 39.2 & 59.9& 42.7& 41.6& 62.1& 45.6& 35.7& 56.8& 38.1& 37.7& 59.3& 40.6\\
        BYOL~\cite{grill2020bootstrap}& 41.2 & 62.6& 45.1& 42.0 &63.3& 46.0& 37.5& 59.4& 40.2& 38.1& 60.1& 41.2\\
        InfoMin Aug.$^{\sharp}$~\cite{tian2020makes}& 40.6 & 60.6& 44.6& 42.5& 62.7& 46.8& 36.7& 57.7& 39.4& 38.4& 59.7& 41.4\\
        SCAN$^{\sharp}$~\cite{wei2020can}& 41.8 & 62.1 & 45.7 &43.2 &63.3 &47.1 & 37.8 & 59.2 & 40.9 &38.8 &60.6 &41.6\\
        DenseCL$^{\sharp}$~\cite{wang2020dense}& 40.3 & 59.9 & 44.3 &- &- &- & 36.4 & 57.0 & 39.2 &- &- &-\\
        PixPro$^{\sharp}$~\cite{xie2020propagate}& 41.4 & 61.6 & 45.4 & - & - & - & - & - & - & - & - & -\\
        \hline
        E-DIY& \textbf{41.5} & \textbf{62.4}& \textbf{45.3}&\textbf{42.5} & \textbf{63.3}& \textbf{46.4}& \textbf{37.5}& \textbf{59.2}& \textbf{40.2}& \textbf{38.3}& \textbf{60.3}& \textbf{41.1}\\
        E-DIY$^{\sharp}$& \textbf{42.1} & \textbf{62.1}& \textbf{46.0}& \textbf{43.4} & \textbf{63.5}& \textbf{47.3}& \textbf{37.9}& \textbf{59.1}& \textbf{40.9}& \textbf{39.1}& \textbf{60.7}& \textbf{42.2}\\
        \hline
    \end{tabular}}
    \label{tab:2}
\end{table*}

\begin{table*}[htb]
    \centering
    \caption{Results of adopting different pre-training models to fine-tune Mask R-CNN with the R50-C4 backbone and different learning schedules on COCO. Following BYOL, E-DIY is trained with 1000 epochs on ImageNet-1K.}
    \scalebox{0.75}{
    \begin{tabular}{l|ccc|ccc||ccc|ccc}
        \hline
        \multirow{3}{*}{Methods}  &  \multicolumn{6}{c||}{COCO$_\mathrm{Det}$} & \multicolumn{6}{c}{ COCO$_\mathrm{InsSeg}$} \\
        \cline{2-13}
        {} &  \multicolumn{3}{c|}{1$\times$ schedule} & \multicolumn{3}{c||}{2$\times$ schedule} & \multicolumn{3}{c|}{1$\times$ schedule}  & \multicolumn{3}{c}{2$\times$ schedule} \\
        \cline{2-13}
        {} &  \multicolumn{1}{c}{AP$^\mathrm{bb}$} & \multicolumn{1}{c}{AP$_{50}^\mathrm{bb}$} & \multicolumn{1}{c|}{AP$_{75}^\mathrm{bb}$} &
        \multicolumn{1}{c}{AP$^\mathrm{bb}$} & \multicolumn{1}{c}{AP$_{50}^\mathrm{bb}$} & \multicolumn{1}{c||}{AP$_{75}^\mathrm{bb}$} &
        \multicolumn{1}{c}{AP$^\mathrm{mk}$} & \multicolumn{1}{c}{AP$_{50}^\mathrm{mk}$} & \multicolumn{1}{c|}{AP$_{75}^\mathrm{mk}$} &
        \multicolumn{1}{c}{AP$^\mathrm{mk}$} & \multicolumn{1}{c}{AP$_{50}^\mathrm{mk}$} & \multicolumn{1}{c}{AP$_{75}^\mathrm{mk}$} \\
        \hline \hline
        FSup-Softmax~\cite{he2020momentum}&  38.2 & 58.2& 41.2 &  40.0 & 59.9& 43.1 & 33.3& 54.7& 35.2 & 34.7& 56.5& 36.9\\
        \hline
        MoCo-v2~\cite{chen2020improved}& 39.5 & 59.1& 42.7 & 41.2 & 61.0& 44.8 & 34.5& 55.8& 36.7& 35.8& 57.6& 38.3\\
        BYOL~\cite{grill2020bootstrap}& 38.6 & 59.2 & 41.6 & 41.2 & 61.4& 44.7 & 33.6 & 55.3 & 35.4& 35.7& 57.8& 38.1\\
        InfoMin Aug.~\cite{tian2020makes}& 39.0 & 58.5& 42.0 & 41.3 & 61.2& 45.0 & 34.1& 55.2& 36.4 & 36.0& 57.9& 38.3\\
        SCAN~\cite{wei2020can} & 40.1 & 60.2 & 43.2 &41.7 &61.7 &45.4 & 34.9 & 56.6 & 37.1 &36.2 &58.3 &38.6\\
        \hline
        E-DIY& \textbf{40.7} & \textbf{60.8}& \textbf{43.8}& \textbf{42.4} & \textbf{62.4}& \textbf{45.7}& \textbf{35.3}& \textbf{57.3}& \textbf{37.7} & \textbf{36.6}& \textbf{58.9}& \textbf{39.1}\\
        \hline
    \end{tabular}}
    \label{tab:3}
\end{table*}

\subsection{Impact on Linear Classification Task}
Though E-DIY achieves consistent improvements on various downstream tasks, one additional question is whether E-DIY could do harm to image classification task for it only requiring limited discriminative information,~\emph{e.g.}, the foreground region, which is similarly contradictory with the motivation of E-DIY. 
Target to answer this question, we further conduct evaluations on linear evaluation task of ImageNet-1K. All of the experimental details are strictly followed with BYOL.
\begin{wraptable}[]{r}{8cm} 
    \centering
    \caption{ Accuracy of adopting different pre-training methods to fine-tune ResNet-50 under linear evaluation task on ImageNet-1K.}
    \begin{tabular}{l|c|c}
        \toprule
        Methods & Top-1 Acc.($\%$) & Top-5 Acc.($\%$) \\
        \hline\hline
        CMC~\cite{tian2019contrastive} & 64.1 & - \\
        SimCLR~\cite{chen2020simple} & 69.3 & 89.0\\
        MoCo-v2~\cite{chen2020improved}& 71.1 & -\\
        InfoMin Aug.~\cite{tian2020makes} & 73.0 & 91.1\\
        BYOL~\cite{grill2020bootstrap} & 74.3 & 91.6\\
        SwAV~\cite{caron2020} & 75.3 & -\\
        \hline
        E-DIY & 73.8 & 91.5\\
        \bottomrule
    \end{tabular}
    \label{tab:4}
\end{wraptable}
As shown in Tab.~\ref{tab:4}, the performance of E-DIY is slightly decreased compared with the baseline BYOL,~\emph{e.g.}, dropping from $74.3\%$ to $73.8\%$ on Top-1 accuracy. Accordingly, we can conclude that though E-DIY targets to extract the naturally existing multi-grained information from as many regions as possible inside each image, it also preserves the most discriminative information well. Therefore, E-DIY is a more reliable pre-training method because of its stable performance on downstream tasks.


\section{Conclusion}
This paper mainly reveals the necessity of avoiding the representation collapse problem in visual pre-training task. Target to maximally explore the multi-grained information inside each image, we present a simple but effective mechanism, called Exploring the Diversity and Invariance in Yourself (E-DIY). By simply pushing the different regions in the same augmented view away and pulling the consistent regions in the different augmented views from the same image together, E-DIY can preserve the diversity of extracted region-level features well and meantime ensure their semantics discrimination ability.
Extensive experiments on various downstream tasks show E-DIY heavily pushes forward the transfer performance of pre-training models. However, this paper exploits the region-level relationship only on the last single feature map. In the future, we will extend our region-level relationship exploring mechanism to multiple layers in the same network for guiding the model to learn more pre-training knowledge.


\bibliographystyle{plain}
\bibliography{neurips_2021}
\end{document}